\UseRawInputEncoding
\documentclass{article} 
\usepackage[final]{colm2025_conference}
\usepackage{ulem}
\usepackage{bm}
\usepackage{amsfonts}
\usepackage{multirow}
\usepackage{mathtools}
\usepackage{microtype}
\usepackage{hyperref}
\usepackage[nameinlink]{cleveref}
\usepackage{url}
\usepackage{booktabs}
\usepackage{graphicx}
\usepackage{lineno}
\usepackage{graphicx}
\usepackage{subcaption}
\usepackage{listings}
\usepackage[T1]{fontenc}
\usepackage[utf8]{inputenc}
\usepackage{makecell} 

\definecolor{darkblue}{rgb}{0, 0, 0.5}
\hypersetup{colorlinks=true, citecolor=darkblue, linkcolor=darkblue, urlcolor=darkblue}

\title{Always Tell Me The Odds:\\Fine-grained Conditional Probability Estimation}

\author{
    \textbf{Liaoyaqi Wang\thanks{Equal contribution.}}{\hspace{.1em}}
    \quad
    \textbf{Zhengping Jiang\footnotemark[1]}{\hspace{.1em}}
    \quad
    \textbf{Anqi Liu}{\hspace{.1em}}
    \quad
    \textbf{Benjamin Van Durme}{\hspace{.1em}}
    \quad
    \vspace{.5em}\\
    Johns Hopkins University
    \vspace{.5em}\\
    \texttt{\{lwang240,zjiang31,aliu74,bvandur1\}@jh.edu}
}

\begin{document}

\ifcolmsubmission
\linenumbers
\fi

\maketitle

\begin{abstract}
We present a state-of-the-art model for fine-grained probability estimation of propositions conditioned on context. Recent advances in large language models (LLMs) have significantly enhanced their reasoning capabilities, particularly on well-defined tasks with complete information. However, LLMs continue to struggle with making accurate and well-calibrated \emph{probabilistic} predictions under uncertainty or partial information. While incorporating uncertainty into model predictions often boosts performance, obtaining reliable estimates of that uncertainty remains understudied. In particular, LLM probability estimates tend to be coarse and biased towards more frequent numbers. Through a combination of human and synthetic data creation and assessment, scaling to larger models, and better supervision, we propose a set of strong and precise probability estimation models. We conduct systematic evaluations across tasks that rely on conditional probability estimation and show that our approach consistently outperforms existing fine-tuned and prompting-based methods by a large margin
\footnote{\url{https://github.com/zipJiang/decoding-based-regression/tree/main}}\footnote{\url{https://huggingface.co/collections/Zhengping/always-tell-me-the-odds-6806b1e01cb76d8c7f3a33ef}}.
\end{abstract}

\section{Introduction}
While large language models (LLMs) have shown remarkable performance in a wide range of well-formulated, clearly solvable reasoning tasks, real-world applications often require them to operate under uncertainty and ambiguity, where purely deductive or deterministic reasoning may fail~\citep{mccarthy1981some}. 
Much of real-world and commonsense knowledge is inherently probabilistic \citep{li2021systematic, moss2018probabilistic, glickman2005probabilistic}. Thus, it is crucial for LLMs to integrate evidence with prior knowledge and to evaluate the plausibility of new information \citep{jaynes2003probability}.
We want LLMs that always (and accurately) tell us the odds.\footnote{\url{https://star-wars-memes.fandom.com/wiki/Never_tell_me_the_odds}}

Various approaches incorporate uncertainty into the LLM reasoning process. For example, a recent line of work proposes to create structured reasoning traces that allow uncertainty to propagate \citep{jung-etal-2022-maieutic, feng2024birdtrustworthybayesianinference, xia2024let,hou2024probabilistic,akyurek2024deductive,sanders2025bonsaiinterpretabletreeadaptivegrounded}. Further, it has been argued that uncertainty-aware evaluation provides better insights into the model capabilities \citep{cheng2024every,jiang2024core,yuan2024probelm}. 
However, these methods typically rely on estimating conditional probabilities over local structures using prompting strategies that are often underexplored and poorly optimized, effectively delegating this task to the LLM without sufficient scrutiny. Approaches typically involve examining the logits of the model \citep{zhao2021calibrate}, which are known to be miscalibrated \citep{jiang2021can,xiong2024can, jiang2022calibrating}, or using verbalized confidence \citep{mielke-etal-2022-reducing, tian-etal-2023-just} which tends to stick to a few common discrete values \citep{razeghi2022impact, cruz2024evaluating,feng2024birdtrustworthybayesianinference}.

Driven by the necessity of accurate, fine-grained probabilistic reasoning, in this work we focus on building strong, LLM-based models for estimating the probability of a textual proposition given a context. Direct evaluation against human annotations can be subjective and noisy; therefore, we first construct an objective and comprehensive evaluation suite by designing subproblems from existing datasets and frameworks that admit an intuitive probabilistic interpretation.

\begin{figure}[t]
\small
\includegraphics[width=\textwidth]{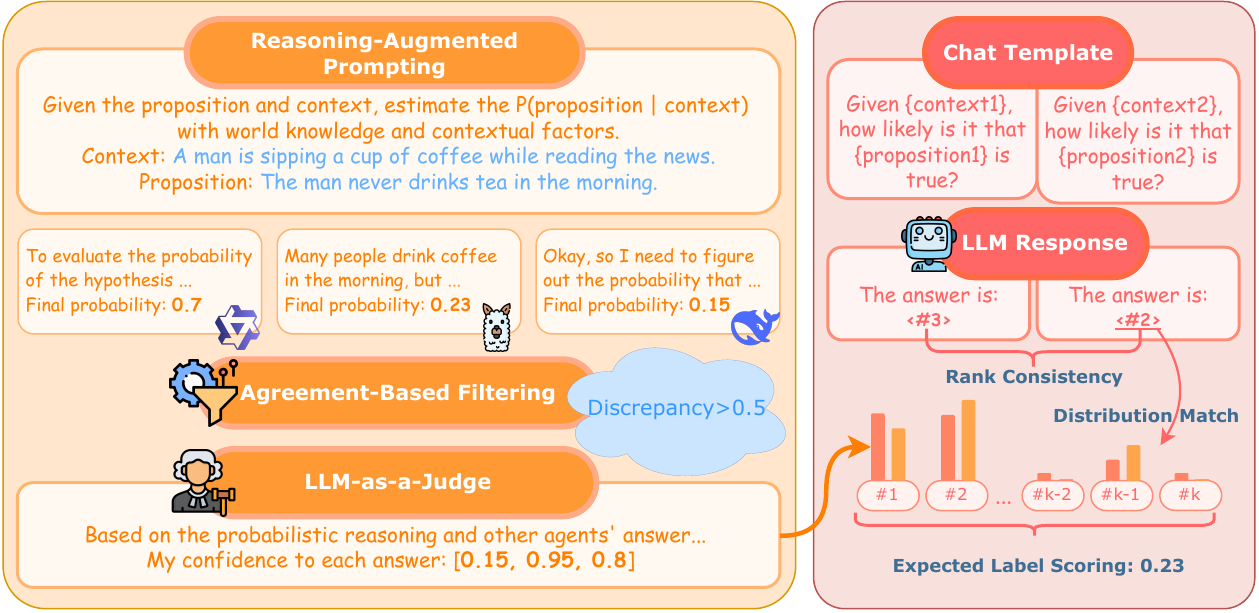}
\caption{We train decoder-based models for fine-grained probability estimation, going beyond human annotations. We rely on two sources of supervision: synthetic (left), and pairwise ranking (right). For synthetic data, we first collect multiple LLMs' probability estimates with reasoning, then group instances by LLM agreement. For those with significant discrepancies, we submit them to another LLM judge, which rate the quality of each reasoning process. These ratings are then used to aggregate the probability estimates into a distribution over possible bins. For pairwise ranking, we use a margin loss to ensure consistency between the pairwise labels and the fine-grained probability estimates -- which are computed via the expected label scoring rule.}
\label{fig::pipeline}
\end{figure}
We therefore propose to train such conditional probability estimation models with (1) a modern LLM backend, (2) much more synthetic data, and (3) diverse training objectives. While most existing approaches rely on regression with smaller encoder-based models to fit human annotation \citep{UNLI, nie-etal-2020-learn}, by leveraging an expected label scoring rule that bridges regression / ranking and calibrated classification \citep{jiang2024addressing}, we propose simple techniques that can provide fine-grained probability estimates with a modern, large decoder-based model. This allows us to better utilize the exceptional language understanding capabilities of pretrained LLMs. Also, our formulation allows natural integration with ensemble distillation using synthetic data as well as training with pairwise likelihood comparison. Due to the challenges in getting high quality data described above, these techniques are crucial to further improve performance.

We also propose a suite of tasks to comprehensively evaluate the performance of such models, ranging from their alignment with human and synthetic annotations, to the consistency of their plausibility rankings with human perception, and their ability to support Bayesian inference and decision-making. Our best model surpasses existing fine-tuned and prompting-based methods across multiple tasks, revealing untapped potential in improving probabilistic reasoning in LLMs through better probability estimation.

\section{Related Work}
\paragraph{Uncertainty in LLMs} Like other modern neural networks \citep{guo2017calibration}, transformer-based language models are often overconfident in their predictions \citep{desai2020calibration, jiang2022calibrating}. This makes LLM bad confidence estimator, while instruction tuning has been reported to further concentrating probability mass \citep{GPT4, hendrycks2020measuring, padmakumar2023does}. To address this, various methods have been proposed to improve uncertainty estimation, such as analyzing logits \citep{zhao2021calibrate}, prompting LLMs to verbalize confidence \citep{mielke-etal-2022-reducing, tian-etal-2023-just}, or probing internal representations \citep{ch-wang-etal-2024-androids}. Of these, verbalized uncertainty has generally been shown to produce the most accurate probability estimation \citep{tian-etal-2023-just}, though it can still be overconfident \citep{tanneru2023quantifyinguncertaintynaturallanguage, xiong2024can, chen2023closelookcalibrationpretrained} and often relies on a small set of common discrete outputs (e.g., 10\%, 75\%) \citep{cruz2024evaluating}.

Part of the problem is that existing resources for training are predominantly hard labeled, making model calibration and its evaluation particularly challenging \citep{kumar2019verified, Nixon_2019_CVPR_Workshops}. Recently, the field has seen growing interest in modeling the complete label distribution \citep{meissner2021embracing, liu2023we, zhou2021distributed, cheng2024every}, and different protocols have been proposed for curating datasets to enable such training and evaluation \citep{nie-etal-2020-learn, UNLI}. However, these datasets tend to be limited in size and are subject to ongoing debate over whether the distributions reflect true aleatoric uncertainty or merely label noise \citep{jiang2022investigating, baan2022stop, wang-etal-2022-capture, weber2024varierr, baan2024interpreting}. 

Thus, to mitigate the limitations of existing resources, we propose training our model with synthetic data and further supervising it with pairwise ranking supervision.

\paragraph{LLM-based Probabilistic Reasoning} Chain-of-thought prompts effectively elicit reasoning in LLMs \citep{wei2022chain}, but do not ensure accurate or calibrated probability estimates in real-world settings \citep{Paruchuri2024WhatAT}. To address this, various methods integrate probability into reasoning. \citet{ozturkler-etal-2023-thinksum} proposed a two-stage framework where LLMs are queried in parallel and aggregated via likelihoods. \citet{feng2024birdtrustworthybayesianinference} and \citet{nafar2024reasoninguncertaintextgenerative} align Bayesian Networks with LLM abductions for decision-making and QA tasks. Others decompose statements into sub-claims and aggregate them using confidence scores~\citep{xia2024let,hou2024probabilistic,cao-etal-2023-probabilistic,jung-etal-2022-maieutic,akyurek2024deductive}. While LLMs can be tuned to mimic Bayesian models \citep{qiu2025bayesian}, such supervision is rarely feasible in practice, limiting prior work to toy examples \citep{wong2023word}. Most existing methods solely rely on aggregating local probability estimates via prompting or probing \citep{jung-etal-2022-maieutic}, which can be suboptimal as discussed above. We show that LLMs can be directly tuned to produce fine-grained, accurate conditional probabilities for structured reasoning.

\paragraph{Decoder-based Regression} Although primarily trained to predict the next token in textual sequences, large language models (LLMs) have been shown to perform effectively as regression models. \citet{vacareanu2024words} demonstrate that LLMs can conduct both linear and non-linear regression in context. In a series of papers, \citet{song2024omnipred} and \citet{song2025decoding} develop techniques for training LLMs to perform numerical regression using only textual representations, achieving arbitrary precision. In this work, to balance efficiency and usability, we propose a hybrid approach that combines coarse-grained probability estimates via textual representations with fine-grained probability refinement through expected label scoring rule aggregation \citep{jiang2024addressing}.

\section{Methodology}
We aim to develop accurate, fine-grained, and broadly applicable models for estimating the conditional probability $P(\text{proposition} \mid \text{context})$, representing the likelihood that a candidate proposition is true or occurs given a textual context.
\hyperref[subsec::objective]{Subsection~\ref*{subsec::objective}} outlines our evaluation suite, which covers intrinsic alignment with labels, consistency of probability rankings, and support for structured reasoning, beyond straightforwardly comparing predictions to noisy human annotations.
\hyperref[synthetic dataset]{Subsection~\ref*{synthetic dataset}} presents our strategy for scaling up training data using LLM-based pseudo-labeling, despite known inaccuracies and biases in LLM probability estimates. Finally, \autoref{subsec::fine-tuning} describes our training procedure for fine-tuning LLM-based models using a combination of human annotations, synthetic probability estimates, and pairwise ranking consistency signals.
\subsection{Objective and Evaluation}
\label{subsec::objective}
What makes a good model for conditional probability estimation? When accurate probabilistic labels are available -- e.g., from census data or controlled Bayesian setups -- evaluation is straightforward. However, such labels are rare in real-world settings, where commonsense reasoning and complex patterns are involved. Human-provided estimates can help \citep{UNLI, nie-etal-2020-learn} but are often subjective and noisy \citep{meissner2021embracing}.

Instead of relying solely on ground-truth alignment, we advocate evaluating models based on their effectiveness in real-world decision-making, where probabilistic reasoning aids belief modeling and evidence integration \citep{feng2024birdtrustworthybayesianinference, xia2024let, qiu2024can}. Inspired by multi-class calibration \citep{zhao2021calibrating}, we hypothesize that better probability estimates should support more human-aligned decisions.

To this end, we propose three task categories: (1) \textbf{Intrinsic}, comparing model estimates with human or LLM labels; (2) \textbf{Comparison}, ranking plausibility among probability estimates; and (3) \textbf{Structural}, assessing uncertainty propagation and decision-making in structured reasoning (\autoref{subsec::dataset}).

\subsection{Synthetic Data Creation}
\label{synthetic dataset}
This section introduces our method for generating pseudo-probabilistic labels for  propositions given textual context, aiming to expand domain and distribution coverage for conditional probability estimation. 

Inspired by the fact that increased inference-time compute improves both performance and confidence \citep{kojima2022large, muennighoff2025s1, jurayj2025your}, we enhance annotation quality by generating multiple LLM roll-outs for each probability estimation~\citep{wang2023selfconsistency}. Prior work highlights a discrepancy between LLMs' generative and evaluative capabilities \citep{gu2024survey, zelikman2022star, kumar2024training}, with models excelling at tasks like judging or verifying outputs. We leverage this by proposing a multi-step annotation aggregation process, where probability estimates are judged based on reasoning quality by another LLM -- yielding consistent performance gains. We explored various ways of improving the pseudo-label quality, including one similar to the confidence ranking approach proposed by \citet{shrivastava2025language} with LLM-based pairwise comparison, which we detailed in \autoref{appendix::other-attempts}. Overall, we find the approach adopted here scales the most efficiently with compute.

\paragraph{Reasoning-Augmented Prompting}
The first step in our annotation process involves eliciting direct probability estimates from LLMs alongside their corresponding reasoning chains. Our prompting strategy instructs LLMs to leverage world knowledge to assess contextual factors before arriving at a final estimation, as incorporating reasoning helps models better approximate human decision-making~\citep{chen-etal-2024-seeing}. This approach encourages models to decompose ambiguous premises into plausible real-world scenarios and estimate the likelihood of each scenario's occurrence~\citep{conf/icml/HouLQAC024}. Intuitively, multiple rounds of annotation using different models and configurations can provide a richer understanding of each data point. We enforce reasoning chains for two main reasons: (1) they tend to enhance performance, and (2) they improve interpretability, enabling more rigorous evaluation and verification~\citep{lightman2023let}, while also serving as useful input for subsequent steps.\footnote{Please refer to \autoref{app_prompt} for the exact prompts we use.}

\paragraph{Agreement-Based Filtering}
After obtaining raw probability scores by directly prompting LLMs, we seek to identify low-quality estimates for further refinement. We observe that estimates with high agreement are more aligned with human annotations in the UNLI validation set~\citep{UNLI}. This is supported by prior findings that ensemble agreement can be a reliable proxy for label quality~\citep{10.5555/3618408.3618712,baek2022agreementontheline}, and that targeting uncertain labels can improve annotation quality~\citep{gligoric2024can}. To quantify confidence, we define \textit{discrepancy} as the difference between the maximum and minimum probability estimates across models. For low-discrepancy samples, we retain the original scores; high-discrepancy samples are flagged for further review.

\paragraph{LLM-as-a-Judge} In the final step, we employ a large language model (LLM) as a judge to adjudicate among candidate reasoning chains~\citep{zheng2023judgingllmasajudgemtbenchchatbot}. Specifically, the LLM evaluates the quality of each reasoning process that leads to a conditional probability estimate~\citep{10.5555/3692070.3692537,chiang2023a}. For reasoning traces produced by reinforcement learning-enhanced models, we first summarize the rationale to extract key steps influencing the probability assessment before presenting it to the judge. This preprocessing step is necessary because such reasoning traces are often lengthy and exhibit complex reasoning patterns~\citep{deepseekai2025deepseekr1incentivizingreasoningcapability}, which may confuse the LLM judge. After analyzing the context, outcome, and candidate reasoning chains, the judge assigns a confidence score between 0 and 1 to reflect the reliability of each chain~\citep{xu2023reasoninglargelanguagemodels}. These confidence scores are then used as supervision signals during model fine-tuning, as discussed below.

\subsection{Model Fine-tuning}
\label{subsec::fine-tuning}
While there already exists a pretraining-based model for subjective probabilities \citep{UNLI}, they are typically small-encoder-based and tuned naively to match human scalar annotation. 
In this section, we describe our training recipe for leveraging modern large-scale decoder-based language models, incorporating calibrated human annotations, utilizing synthetic labels for domain generalization, and integrating pairwise ranking consistency signals to enhance model performance.

\paragraph{Decoder-based Regression} refers to the practice of training decoder-only language models to perform regression by outputting textual representations of numeric values~\citep{song2025decoding}. While LLM embeddings have been used as features for regression tasks, prior studies report unclear scaling trends~\citep{tang2024understanding}, as these embeddings mostly capture semantic similarity rather than supporting precise numerical prediction~\citep{devlin2019bert, li2020sentence}. Besides, we prefer textual-based outputs as users will be able to directly get the outputs from common LLM services without additional processing~\citep{kwon2023efficient}.

\begin{figure}[htbp]
  \centering
  \includegraphics[width=\textwidth]{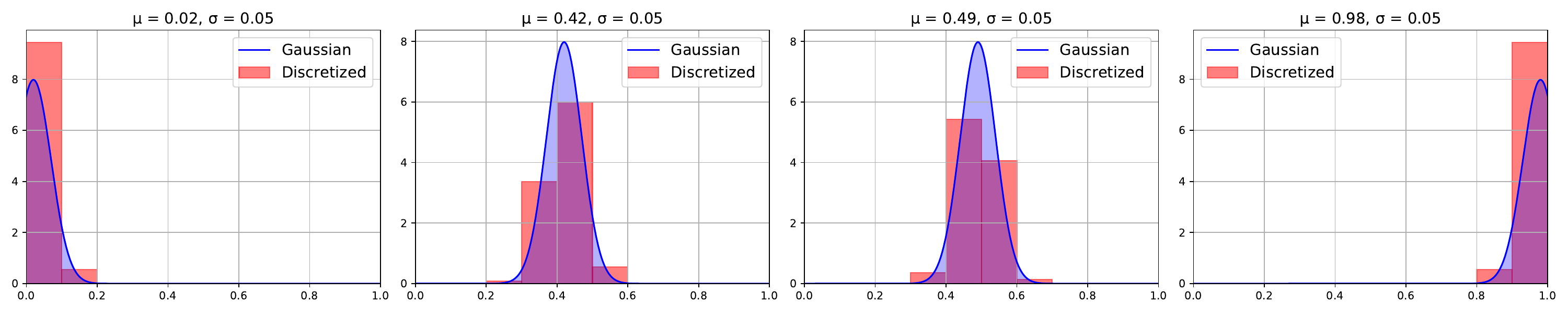}
  \caption{Illustration of our distribution quantization process. Notice that how the quantization preserves fine-grained label ordering and allows for better discrimination of targets that fall in the same bin.}
  \label{fig::discretization}
\end{figure}

Specifically, we split the interval $[0, 1]$ into $N$ bins of equal width $\{b_0, b_1, \dots, b_{n - 1}\}$,  and assign each bin $b_j$ a unique special token $t_j$. 
For an instance $(x_i, y_i) \in \mathcal{D}$ (where $\mathcal{D}$ denotes the dataset), we tune the decoder-only model to predict the token $t_j$ corresponding to the bin $b_j$ that contains the target probability $y_i$. However, this conversion $y \mapsto t$ is lossy, resulting in a coarse prediction. To recover fine-grained scalar prediction, we consider taking the expectation over all possible token predictions. Suppose we have a scoring function $f$ that converts each bin into a scalar $f \coloneqq \mathcal{B} \rightarrow \mathbb{R}^+$, we construct our fine prediction as
\begin{align*}
    \hat{y} = \sum_{j = 0}^{N - 1} f(b_j) \cdot p(t_j \mid x_i).
\end{align*}
This is called the expected label scoring rule, introduced by \citet{jiang2024addressing}, shows that both the mean absolute error and the ranking risk regret lower-bound the calibration error of the underlying classifier. In our case, this implies that the better the model approximates the true token distribution, the more accurately it captures the underlying conditional probability.
Inspired by their work, we convert the ground truth $y_i$ to a Gaussian distribution centered at $y_i$ with a small fixed variance $\sigma^2$. We then quantize this distribution into a discrete distribution with $N$ fixed supports at $\{f(b_0), \dots, f(b_{N - 1})\}$, as \citet{jiang2024addressing} has shown that the Wasserstein-2 distance minimizing quantization is given by
\begin{align*}
  q(t_j|x) = F\Big(\frac{f(b_{j + 1}) + f(b_{j})}{2}\Big) - F\Big(\frac{f(b_{j}) + f(b_{j - 1})}{2}\Big),
\end{align*}
where $F(\cdot)$ is the CDF function of the Gaussian distribution $\mathcal{N}(y_i, \sigma^2)$.
\autoref{fig::discretization} shows some example quantization. We then use the quantized distribution as the target for our model training. This allows us to further discriminate targets that fall in the same bin. While prior work has been shown that enforcing distribution over a set of sequences can be achieved through sampling and reweighting \citep{zhangforcing}, our regression setting only requires a single decoding step. This enables us to directly optimize the model using a forward KL-divergence loss:
\begin{equation*}
  \mathcal{L}_{\text{Direct}}(\mathcal{D}) = \mathbb{E}_{(x, y) \sim \mathcal{D}}D_{\text{KL}}\Big(Q(t|x) ||  P_{\theta}(t|x)\Big).
\end{equation*}
We use forward KL-divergence as we observe more stable training, and it seems that the theoretical mode-seeking behavior of reverse-KL does not always hold for LLMs \citep{wu2025rethinking}.
\paragraph{Utilizing Synthetic Data} While our approach to regression allows us to convert any target scalar label $y$ into a distribution over {bin-associated} tokens, our synthetic data is annotated with multiple LLMs to provide a more reliable estimate of the true conditional probability. On each synthetic data point $x \in \mathcal{X}$, our synthetic data creation process generates $K$ probability estimates $\{y^0, \dots, y^{k - 1}\}$, paired with $K$ confidence judgments $\{c^{0}, \dots, c^{k -1}\}$. As through the previous discussion, each of the score can be mapped to a distribution over tokens. To increase sharpness of the overall prediction, we construct a mixture distribution over $\{Q^0(t|x), \dots, Q^{k - 1}(t|x)\}$, where each component is weighted by a power-normalized confidence score:
\begin{equation*}
  Q(t|x) = \sum_{k = 0}^{K - 1} \pi^k \cdot Q^k(t|x),\ \text{where}\ \pi^k = \frac{(c^k)^{\alpha}}{\sum_{j = 0}^{K - 1} (c^j)^{\alpha}}.
\end{equation*}

\paragraph{Rank Consistency Training} As ground truth probability estimates are only available on very limited data points, we can still leverage other forms of supervision to improve model performance. Similar to how we construct the evaluation suite from defeasible NLI \citep{rudinger-etal-2020-thinking} and choice of plausible alternatives \citep{10.1007/11736790_9,zellers-etal-2019-hellaswag}, we can supervise rank-consistency using pairwise ranking labels. Given two instance $(x_1, y_1)$, $(x_2, y_2)$, we use the pairwise margin-loss \citep{weston1999support} to enforce the order consistency:
\begin{equation*}
  \mathcal{L}_{\text{Rank}}(\mathcal{D}_{\text{Pairwise}}) = \mathbb{E}_{\mathcal{D}_{\text{Pairwise}}} \max\{0, \delta - \text{sgn}(y_1 - y_2) \cdot (\hat{y}_1 - \hat{y}_2)\},
\end{equation*}
as it has been shown to lead to less over-confident probability estimates \citep{li2019learning}. Thus the final loss becomes
\begin{equation*}
  \mathcal{L}(\mathcal{D}) = \mathcal{L}_{\text{Direct}}(\mathcal{D}_{\text{Human}}) + \beta_1 \mathcal{L}_{\text{Direct}}(\mathcal{D}_{\text{Synthetic}}) + \beta_2 \mathcal{L}_{\text{Rank}}(\mathcal{D}_{\text{Pairwise}}).
\end{equation*}

Notice that while \citet{jiang2024addressing} empirically show that optimizing for rank-consistency improves the calibration of the underlying classifier, their theoretical results depend on the assumption that the predictive distribution is single-peaked. For our case this does not always hold, and in fact our gradient analysis in \autoref{appendix::gradient} shows that the margin-loss as well as expected-scoring-rule-based regression loss alone can be under-specified, and does not guarantee that greedy decoding will yield a reasonable approximation of the target conditional probability. Therefore, we apply rank-consistency training only when the pairwise labels are available, to improve correspondence of the probability estimates.

\section{Experiment Setup}
\subsection{Dataset}
\label{subsec::dataset}

\paragraph{Training} We construct our training dataset from a mixture of human-annotated, synthetic, and pairwise ranking data. As our tasks are closely related to Natural Language Inference (NLI), we heavily rely on NLI datasets for training. We use the following datasets: \textbf{UNLI} \citep{UNLI}, which contains high-quality subjective probabilistic relabeling of a subset of SNLI \citep{bowman-etal-2015-large}; \textbf{ANLI} \citep{nie-etal-2020-adversarial}, which contains adversarially generated NLI examples that are often longer and involve more challenging reasoning patterns \citep{williams2020anlizing}; and \textbf{WANLI} \citep{liu-etal-2022-wanli}, which consists of automatically generated NLI pairs seeded from challenging instances in MultiNLI \citep{williams2018broad}, identified through data maps \citep{swayamdipta-etal-2020-dataset}. 

Since UNLI applies a logistic transformation to its annotation scale, we invert that transformation to make our model more sensitive to probability differences near 0 and 1. As ANLI and WANLI do not provide probabilistic labels, we use our proposed synthetic data annotation pipeline, detailed in \autoref{synthetic dataset}, to generate pseudo-labels for these datasets.

Additionally, we leverage \textbf{$\delta$-NLI} \citep{rudinger-etal-2020-thinking} and \textbf{HellaSwag} \citep{zellers-etal-2019-hellaswag} for rank-consistency training. For $\delta$-NLI, we train our models' probability estimates to be consistent with the direction of defeasible updates, while for HellaSwag, we train the models to assign the highest probability to the most plausible completion. During training, we upsample UNLI and $\delta$-NLI by up to 4 times~\citep{lee-etal-2022-deduplicating,li2025upsampleupweightbalancedtrainin}.

\paragraph{Intrinsic Evaluation} Besides evaluating our models on the \textbf{UNLI} dataset, we also sample from \textbf{EntailmentBank} \citep{dalvi-etal-2021-explaining} and \textbf{e-CARE} \citep{du-etal-2022-e} to directly evaluate the quality of our probability estimates. While EntailmentBank trees are predominantly entailment-focused, we rewrite hypotheses to introduce probabilistic uncertainty by removing modality, increasing vagueness, generating alternatives, perform existential instantiation, and generate abductions to convert the task into a probabilistic reasoning challenge.
To evaluate the quality of probability estimations on other domains, we follow \textbf{GNLI} \citep{hosseini2024synthetic} to generate synthetic NLI pairs in more distant domains such as fans forum, blog post, medical texts, and others.
We also cast the scalar annotation \citep{jiang2024addressing} on Circa \citep{louis-etal-2020-id} to the probability of an ambiguous response being affirmative. For all intrinsic evaluation, we report Spearman correlation.

\paragraph{Comparison Evaluation} We evaluate accuracy of plausibility ranking on \textbf{$\delta$-NLI}, \textbf{HellaSwag} and \textbf{COPA} \citep{roemmele2011choice}. Specifically, for COPA, we rank the two choices based on the conditional probability of the effect given the cause, regardless of the question type. For each instance, our model is required to predict conditional probability for each of the alternatives independently, which makes the task more challenging than the original multiple-choice setting. We report the accuracy of the model's top-1 ranking.

\paragraph{Structural Evaluation} 
We evaluate our model on reasoning traces from two reasoning frameworks. The first is \textbf{Maieutic Prompting} \citep{jung-etal-2022-maieutic}, which iteratively expands a maieutic tree using abductive and recursive explanations, solving for the most consistent truth value assignment. While the original formulation incorporates multiple methods to estimate uncertainty and applies post-hoc filtering, we make the task more challenging by relying solely on our model to estimate all required conditional probabilities. Specifically, we estimate both \textit{belief} and \textit{consistency}, and we remove the NLI filtering constraint to further increase difficulty.

The second framework is \textbf{BIRD} \citep{feng2024birdtrustworthybayesianinference}, which performs Bayesian inference through LLM-generated abductions. Given a BIRD trace, we use our model to score both the conditional probabilities of each factor condition given the context, and the probabilities of each final outcome given those conditions. Unlike the original BIRD method, we do not apply any optimization for consistency, aside from renormalizing the aggregated probabilities over final outcomes.

For Maieutic Prompting, we evaluate on publicly available traces from \textbf{Com2Sense} \citep{singh-etal-2021-com2sense}, \textbf{CREAK} \citep{2021_5737c6ec}, and \textbf{CSQA2} \citep{talmor2021commonsenseqa}—all binary QA benchmarks focused on commonsense reasoning and fact verification. For BIRD, we use their released dataset, including additional sentence generation for the comparison subset of \textbf{Com2Sense}, and evaluate on \textbf{Today} \citep{feng2023generic} for temporal reasoning. We report overall accuracy across all datasets. Example traces and scoring details can be found in \autoref{appendix::mermaid}.

\subsection{Model and Tuning Details} 

We use four open-source models to generate synthetic datasets: Qwen2.5-32B-Instruct and QwQ-32B~\citep{qwen2025qwen25technicalreport}, DeepSeek-R1-Distill-Qwen-32B~\citep{deepseekai2025deepseekr1incentivizingreasoningcapability}, and Llama-3.3-70B-Instruct~\citep{grattafiori2024llama}. For all models except QwQ, we apply greedy decoding; for QwQ, we follow the official recommendation of using temperature $= 0.6$ and MinP $= 0$ to mitigate repetition.\footnote{\url{https://huggingface.co/Qwen/QwQ-32B}}

For fine-tuning, we consider three base models: Llama-3-8B-Instruct, Qwen2.5-7B-Instruct, and Qwen2.5-14B-Instruct. We apply LoRA~\citep{hu2022lora} with $r = 16$ and $\alpha = 32$, using a learning rate of $2 \times 10^{-5}$ and a linear learning rate scheduler. For each configuration, we perform a hyperparameter sweep over the number of levels $\in \{10, 20, 100\}$, $\beta_2 \in \{1, 10, 100\}$, and $\sigma \in \{0.01, 0.05, 0.1\}$.

{We evaluate our proposed models against a diverse set of baselines to ensure a comprehensive comparison. These baselines are: 1) Encoder Model: The RoBERTa-L
  model~\citep{UNLI}, a fine-tuned encoder using UNLI dataset. 2) Zero-Shot LLMs: We prompt both a proprietary model (GPT-4o~\citep{GPT4}) and an open-source model (DeepSeek-R1-Distill-Qwen-32B~\citep{deepseekai2025deepseekr1incentivizingreasoningcapability}) in a zero-shot inference setting. Due to the slow inference speed of GPT-4o on structural reasoning tasks with the Maieutic Prompting method, we employ stratified sampling to create a representative mini-batch for its evaluation. 3) Probe Method: We adopt the true/false probing technique from~\citet{tian-etal-2023-just}. This involves extracting probabilities from the "true" and "false" token logits of the Qwen2.5-14B-Instruct model, which are then calibrated with a general temperature scaling method~\citep{pmlr-v235-shen24c}.}

\section{Results and Discussion}
\label{sec::results-discussions}
Our comprehensive evaluation result is shown in \autoref{tab::overall-results}. Overall, we observe noticeable improvements over previous generation models for subjective probability estimation \citep{UNLI}, as well as over 0-shot prompting approaches that are widely considered. We discuss particular observations below.
\begin{table}[htbp]
\begin{center}
 \resizebox{\linewidth}{!}{
\begin{tabular}{rr|ccccccccc}
\toprule
\multirow{2}{*}{\textbf{Type}}& \multirow{2}{*}{\textbf{Tasks}} & \textbf{Encoder} & \multicolumn{2}{c}{\textbf{0-Shot}} & \textbf{Probe} & \multicolumn{5}{c}{\textbf{Ours}} \\
& & RoBERTa-L & \raisebox{-2pt}{\includegraphics[height=1.5\fontcharht\font`\A]{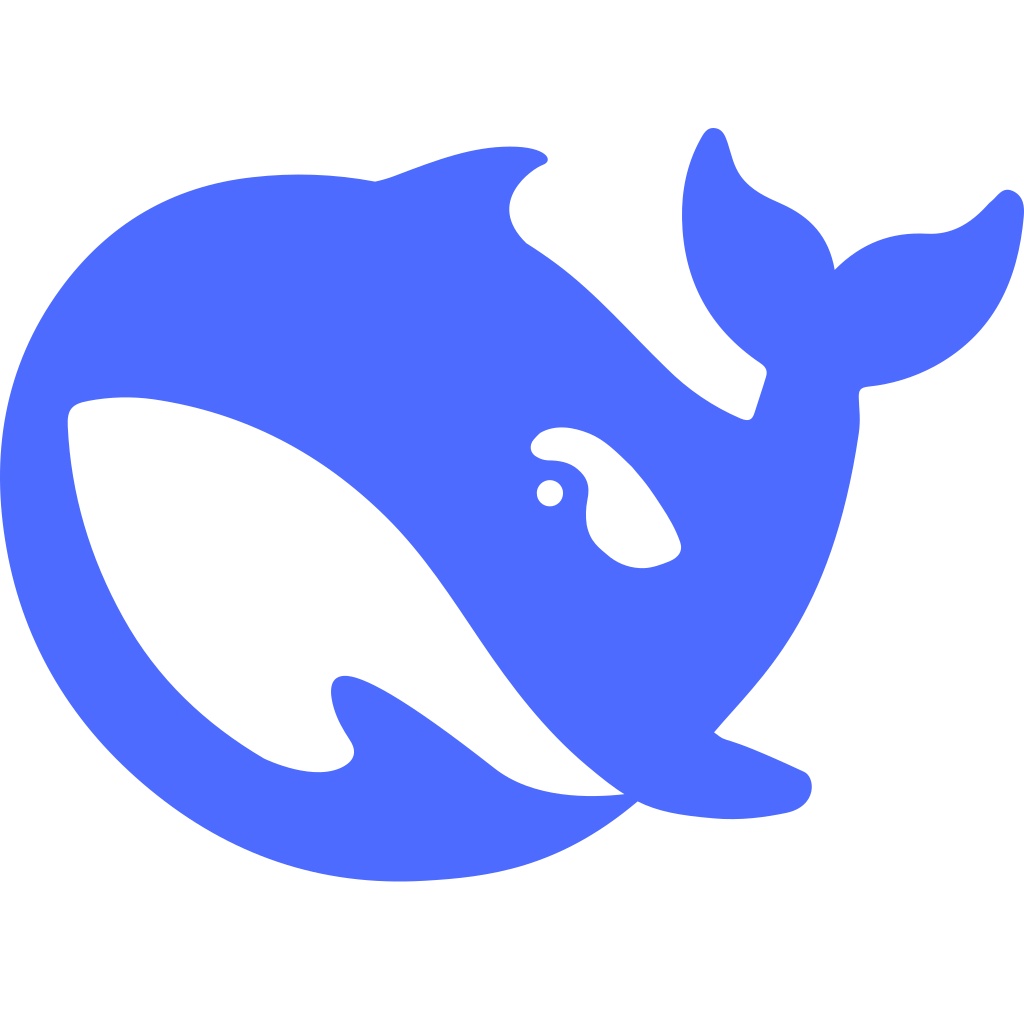}} & \raisebox{-2pt}{\includegraphics[height=1.5\fontcharht\font`\A]{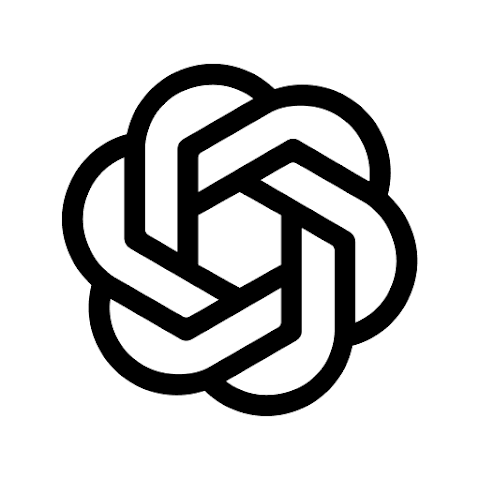}} & \raisebox{-1pt}{\includegraphics[height=1.2\fontcharht\font`\A]{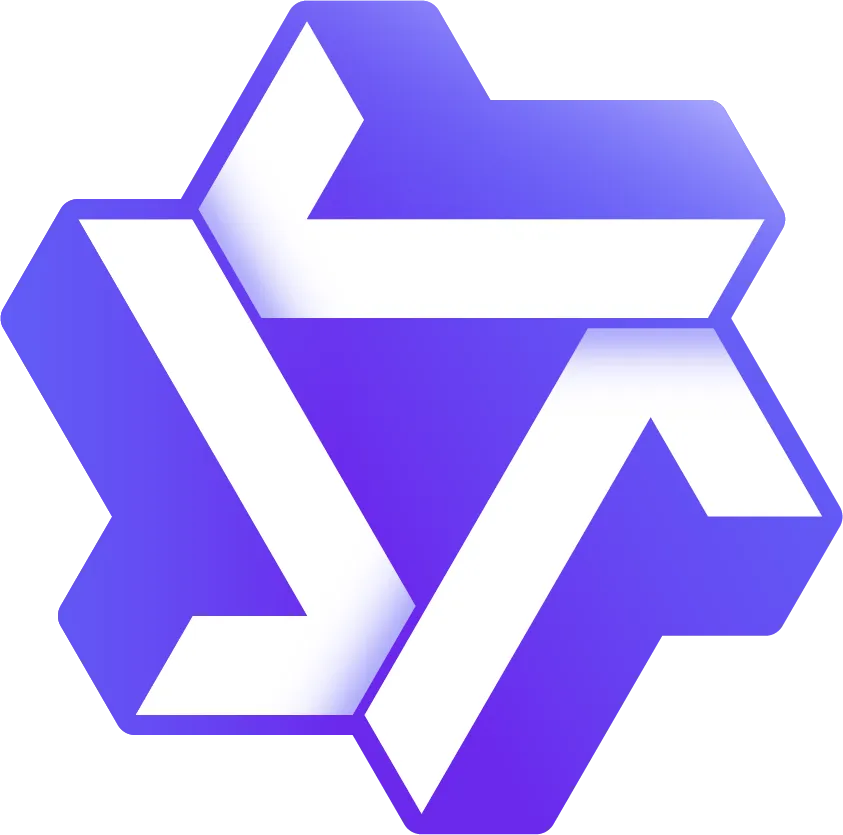}}-14B & \raisebox{-2pt}{\includegraphics[height=1.5\fontcharht\font`\A]{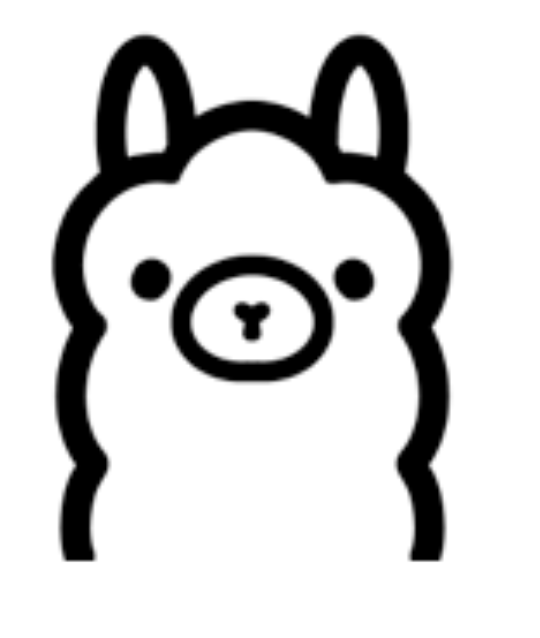}}-8B & \raisebox{-1pt}{\includegraphics[height=1.2\fontcharht\font`\A]{images/qwen.png}}-7B & \raisebox{-1pt}{\includegraphics[height=1.2\fontcharht\font`\A]{images/qwen.png}}-14B & +Syn & +Syn+R \\
\midrule
\multirow{5}{*}{\rotatebox{90}{Intrinsic}} & UNLI & .707 & .629 & .699 & .681& .804 & .802 & \textbf{.813} & \underline{.812} & \underline{.812} \\
& circa & .430 & .663 & \underline{.734} &.553 & .474 & .544 & .536  & .564 & \textbf{.747} \\
& GNLI & .586 & .755 & .796 &.843 & .789 & .811 & .814 & \textbf{.838} & \underline{.820} \\
& EntailmentBank & .503 & .783 & .760 & .558& .659 & .687 & .735 & \textbf{.789} & \underline{.787} \\
& e-CARE & .738 & .871 & \underline{.898} &.856 & .855 & .870 & .888 & .884 & \textbf{.905} \\
\midrule
\multirow{4}{*}{\rotatebox{90}{Comp.}} & $\delta$-SNLI & 77.9 & 77.2 & 75.3 &81.3 & 83.3 & 84.4 & 85.1 & \underline{86.0} & \textbf{88.9} \\
& $\delta$-ATOMIC & 75.0 & 69.1 & 70.7 & 74.7& 78.9 & 78.6 & 80.1 & \underline{81.2} & \textbf{87.6} \\
& COPA & 83.0 & 87.7 & 81.0 &86.8 & 85.1 & 87.2 & 86.5 & \underline{87.9} & \textbf{89.3} \\
& HellaSwag & 42.0 & 57.8 & 75.4 &74.9 & 67.2 & 70.2 & 75.3 & \underline{75.5} & \textbf{95.7} \\
\midrule
\multirow{5}{*}{\rotatebox{90}{Structural}} & C2S-M & 50.6 & 68.5 & \textbf{77.9} &49.4 & 65.5 & 68.4 & 73.5 & {75.2} & \underline{75.6} \\
& CREAK-M & 62.5 & 68.0 & 81.1 &42.1 & 83.2 & 82.8 & 84.8 & \underline{85.6} & \textbf{86.5} \\
& CSQA2-M & 49.5 & 55.6 & \underline{71.4} & 48.5& 60.4 & 63.0 & 67.9 & {70.4} & \textbf{72.0} \\
& C2S-Sent-B & 65.0 & 73.0 & \underline{76.8} &51.3 & 72.7 & 76.3 & 71.3 & 73.3 & \textbf{89.8} \\
& TODAY-B & 56.0 & 64.0 & 63.0 &65.0 & 65.0 & 64.0 & \textbf{68.0} & \underline{67.0} & 66.3 \\
\bottomrule
\end{tabular}}
\caption{Evaluation of our models on the tasks listed out in \autoref{subsec::dataset}. \raisebox{-2pt}{\includegraphics[height=2ex]{images/deepseek-color.png}} denotes the DeepSeek-R1-Distill-Qwen-32B \citep{deepseekai2025deepseekr1incentivizingreasoningcapability} and \raisebox{-2pt}{\includegraphics[height=2ex]{images/gpt-4o.png}} GPT-4o \citep{GPT4}. \raisebox{-2pt}{\includegraphics[height=2ex]{images/llama3.3.png}} corresponds to Llama-3-Instruct series \citep{grattafiori2024llama}, \raisebox{-1pt}{\includegraphics[height=2ex]{images/qwen.png}} the Qwen2.5-Instruct series \citep{qwen2025qwen25technicalreport}. \textbf{+Syn} indicates Qwen2.5-14B-Instruct augmented with synthetic data, \textbf{+R} indicates the same model with rank consistency training. -M and -B suffixes denote the Maieutic Prompting and Bird frameworks respectively. Best results are in \textbf{Bold}, second-best results are \underline{underlined}. 
}
\label{tab::overall-results}
\end{center}
\end{table}
\paragraph{Model performance improves with scale.} Even when trained on the same dataset as prior work \citep{UNLI}, our models built on top of modern LLM backend consistently outperform both smaller encoder-based approaches and zero-shot prompting baselines. Notably, \citet{UNLI} report an aggregated human performance of Spearman's $\rho = 0.727$, whereas all of our models exceed this benchmark by a substantial margin. These results suggest that large language models, with their extensive world knowledge, can be effectively leveraged for probabilistic inference.

\paragraph{Synthetic data enhances domain generalization.} Comparing the results in column \textbf{+Syn} with those in column \raisebox{-1pt}{\includegraphics[height=1.2\fontcharht\font`\A]{images/qwen.png}}-14B, we find that incorporating synthetic data generally improves performance. These improvements are especially notable in probability estimation for more challenging reasoning tasks (e.g., EntailmentBank) and in generalization to distant domains (e.g., GNLI, Circa). We hypothesize that this is due to the broader range of inference types covered by ANLI and WANLI, as well as their longer contexts, which may help the model better understand complex scenarios and integrate multiple pieces of evidence.

\paragraph{Rank-consistency training facilitates better decision making.} The results in column \textbf{+Syn+R} clearly demonstrate that rank-consistency training further improves performance across multiple tasks, particularly in plausibility ranking and structured reasoning. While this outcome is intuitive, it is noteworthy that on some tasks—such as $\delta$-SNLI and C2S-Sent-B our models perform competitively or even outperform previous results, of larger models or systems employing more complex processing pipelines \citep{srikanth2025nli,feng2024birdtrustworthybayesianinference}. Although the improvements with Maieutic Prompting are less pronounced, this is likely due to its lack of a fully probabilistic interpretation \citep{jung-etal-2022-maieutic}.
\begin{figure}[htp]
  \centering
  \begin{minipage}[t]{0.31\textwidth}
    \centering
    \includegraphics[width=\linewidth]{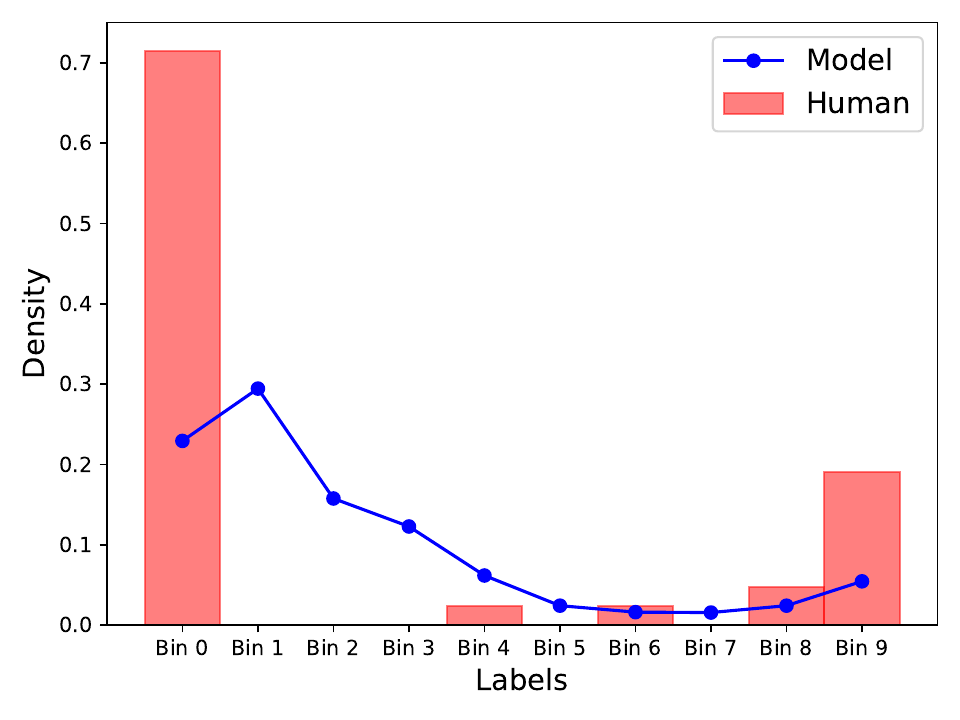}
    \caption*{\tiny\textbf{P}: Ruth's 1927 single season record of 60 home runs stood unsurpassed until Roger Maris hit 61 in 1961. $\leadsto$ \textbf{H}: Babe Ruth hit 60 home runs in his lifetime.}
  \end{minipage}%
  \hspace{1em}
  \begin{minipage}[t]{0.31\textwidth}
    \centering
    \includegraphics[width=\linewidth]{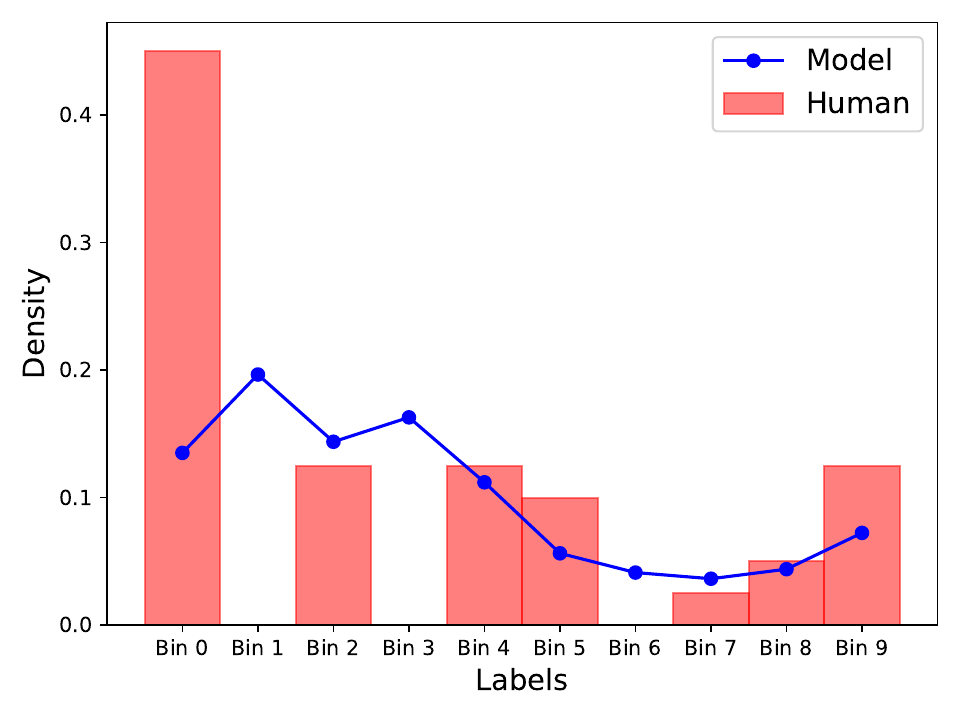}
    \caption*{\tiny\textbf{P}: If no settlement is reached, a divided Cyprus will join the European Union on May 1, 2004. $\leadsto$ \textbf{H}: Cyprus was divided into 2 parts on May 1, 2004.}
  \end{minipage}%
  \hspace{1em}
  \begin{minipage}[t]{0.31\textwidth}
    \centering
    \includegraphics[width=\linewidth]{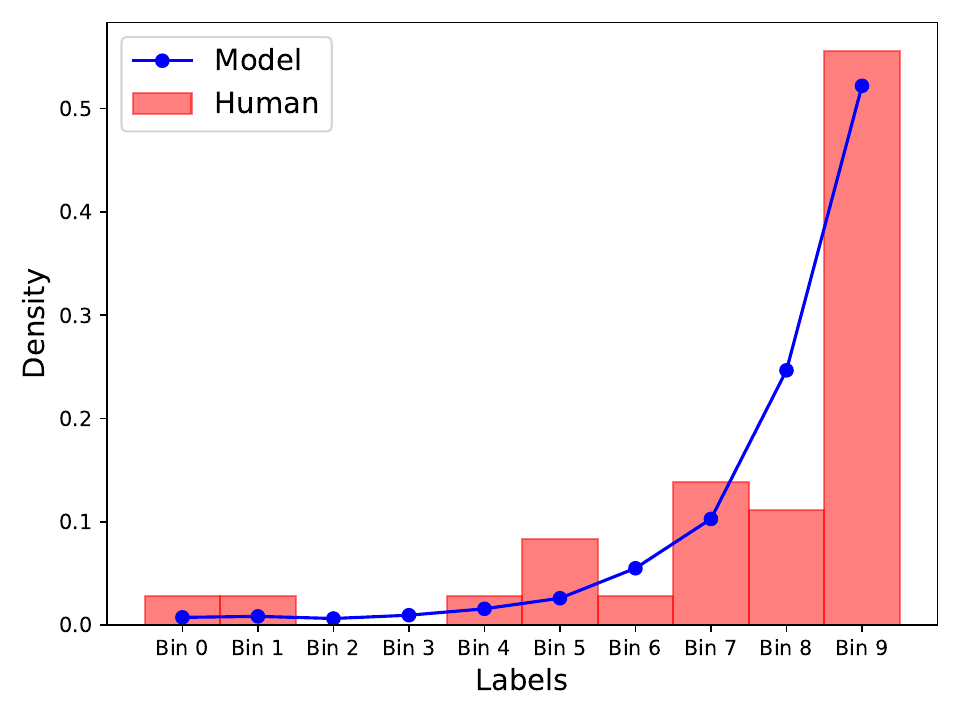}
    \caption*{\tiny\textbf{P}: Man in green vest directing traffic in snowy conditions, pedestrians standing on the sidelines. $\leadsto$ \textbf{H}: A person remains in the conditions.}
  \end{minipage}
  \caption{Comparison of token-level distribution between our model and human label distribution. The x-axis represents the probability bins, while the y-axis indicates the probability density. The distribution learned by our model reflects intrinsic human disagreements.}
  \label{fig::uncertainty}
\end{figure}
\paragraph{Our model captures some level of human uncertainty.} An additional benefit of our model is that, even without explicit training on human label distributions, it still exhibits human-like uncertainty on many data points, as shown in \autoref{fig::uncertainty}. To show this, we compare the token-level distribution of our model with the distributed scalar judgments collected by \citet{pavlick-kwiatkowski-2019-inherent}. We observe that our model produces multi-modal distributions when human opinion divides, and agree with human judgment when the opinion is more consistent. {A more detailed analysis of our model's performance on the ChaosNLI~\citep{nie-etal-2020-learn} and ProtoQA~\citep{boratko-etal-2020-protoqa} datasets is provided in \autoref{appendix::align_with_human}.}

\section{Conclusion}
In this work, we present a series of strong and precise probability estimation models, developed through a novel pipeline that combines LLM-based synthetic data generation with decoder-based regression fine-tuning. Our models consistently outperform previous generation approaches for similar tasks, as well as commonly used zero-shot prompting methods with strong LLMs. Through comprehensive evaluation, we demonstrate that improving local probabilistic estimation can significantly enhance the performance of complex probabilistic reasoning systems. We hope our work inspires future advancements in building efficient, accurate models for real-world probability estimation and, more broadly, in developing general and versatile probabilistic reasoning systems.

\section{Acknowledgments}
This work is supported by U.S. National Science Foundation under grant No. 2204926, ONR grant (N0001424-1-2089) and Defense Advance Research Projects Agency (DARPA) under Contract No. HR001125C0304. Any opinions, findings and conclusions or recommendations expressed in this material are those of the author(s) and
do not necessarily reflect the views of the National Science Foundation and DARPA.

\bibliography{colm2025_conference}
\bibliographystyle{colm2025_conference}
\newpage
\appendix
\section{The Other Probability Estimation Attempt}
\label{appendix::other-attempts}
In this section, we introduce an alternative approach to probability estimation by scaling compute for synthetic data creation. Inspired by the preference data used in Reinforcement Learning from Human Feedback, we hypothesize that pairwise comparison is cognitively easier and more reliable than direct scoring on probability estimation task~\citep{ziegler2020finetuninglanguagemodelshuman}. Our method consists of three main steps: \textbf{pairwise comparison, results aggregation and score mapping}. 
We describe each component in detail below.

\subsection{Pairwise Comparison}

The core of this probability synthesis pipeline is the pairwise comparison of NLI samples using LLMs. Given two given NLI samples \( A \) and \( B \), each consisting of a premise and hypothesis, the LLM is prompted to determine which sample has a higher probability of hypothesis conditioned on the premise. We adopt the prompt format logic introduced by~\citet{qin2024lampo}, for example: 

\textit{"Compare the probabilities of two NLI samples. Passage A: \{A\}, Passage B: \{B\}."} 

For specific prompts, please refer to Appendix~\ref{app_prompt}. To mitigate potential positional biases~\citep{zheng2023judgingllmasajudgemtbenchchatbot,li2024split,shi2024judgingjudgessystematicstudy}, where an LLM might prefer either the first or second sample regardless
of its content, we alternate the order of \( A \) and \( B \) in the prompt. A preference for $A$ over $B$, denoted \( p(A \succ B) \) is considered valid only if the model consistently ranks $A$ higher across both orderings; otherwise, the comparison is treated as a draw~\citep{qin2024lampo}.

\subsection{Results Aggregation}

To construct a global ranking of the test dataset, we aggregate pairwise comparison results using the TrueSkill framework~\citep{trueskill}. In this setup, each NLI sample is modeled as a “player” with a Gaussian skill distribution characterized by mean \( \mu \) and variance \( \sigma^2 \).
For every comparison between two samples, TrueSkill updates their skill distributions accordingly.

The computational cost of this aggregation is efficient, operating near the theoretical minimum of \( n\log(n) \) comparisons for a dataset of size \( n \)~\cite{trueskill}. 
To further reduce comparison overhead, we adopt a binning strategy. The test set is divided into \( m \) bins based on scores from direct prompting, e.g., with \( m=5 \), bins correspond to levels like "impossible","technically possible", "plausible", "likely" and "very likely". During ranking, comparisons are limited to samples within the same bin, effectively reducing comparison times to converge, as previous researches show that adaptive selection of comparisons can reduce the number of required assessments while maintaining ranking accuracy~\citep{10.5555/2986459.2986709,yona2022active}. 
We iterative updates each sample’s Gaussian distribution until its variance \( \sigma \) falls below a predefined threshold \( \delta \), indicating a stable estimate.

\subsection{Score Mapping}
\label{Score Mapping}
In this final step, we map the relative rankings of NLI samples into scalar probability scores. While TrueSkill provides meaningful relative rankings, its outputs are not directly interpretable as probabilities. To bridge this gap, we adopt the Plackett–Luce approximation~\citep{af5079a1-8ca5-3727-a405-0a82390327b7}. Specifically, we use the ratio $\mu/\sigma$ from each sample’s distribution as a proxy "score", and apply a softmax transformation to obtain a calibrated probability distribution.

\subsection{Experiment Setup}
We conduct our experiments on a subset of the UNLI test split~\citep{UNLI}, focusing exclusively on NLI examples labeled as ``neutral''. This is because samples labeled as "entailment" or "contradiction" typically have ground-truth probabilities close to 1 or 0, respectively, whereas our interest lies in distinguishing uncertainty in the middle range. Since we are working with a small dataset, we only adopt the basic strategy to randomly sample compared NLI samples and update their scores using the TrueSkill framework. The random seed is fixed at 42 to ensure the same comparison process for different models.
We initialize each player's mean randomly between 0 and 1 and set the initial variance $\sigma=3$, where a larger variance reflects higher uncertainty in the probability estimate. 

We evaluate performance using two primary metrics:
\textbf{Spearman correlation} between human annotations and the synthesized probability scores, which reflects overall ranking quality;
Pairwise comparison accuracy (\textbf{Compare Acc}), which measures agreement between model comparisons and the ground-truth ordering.
We also track how the correlation evolves as the number of comparison iterations increases.

For the comparison model, we test the performance on DeepSeek-R1-Distill-Qwen-32B~\citep{deepseekai2025deepseekr1incentivizingreasoningcapability} and QwQ-32B~\citep{qwen2025qwen25technicalreport}.
\begin{table}[h]
\begin{center}
\begin{tabular}{lll}
\toprule
\multicolumn{1}{c}{\bf Method}  &\multicolumn{1}{c}{\bf Correlation}&\multicolumn{1}{c}{\bf Compare Acc} \\
\midrule
\raisebox{-2pt}{\includegraphics[height=1.5\fontcharht\font`\A]{images/deepseek-color.png} Pairwise Compare}       &0.6420 & 0.7562 \\
\midrule
{\includegraphics[height=1.5\fontcharht\font`\A]{images/deepseek-color.png} Direct Prompting}  &0.6872 & / \\
\midrule
{\includegraphics[height=1.5\fontcharht\font`\A]{images/qwen.png} Pairwise Compare} &0.6991 & 0.7655 \\
\midrule
{\includegraphics[height=1.5\fontcharht\font`\A]{images/qwen.png} Direct Prompting}  &0.7414 & / \\
\bottomrule
\end{tabular}
\end{center}
\caption{Pairwise Compare Results}\label{pairwise_compare_tab}
\label{tab::compare_result}
\end{table}
\begin{figure}
    \centering
    \includegraphics[width=0.5\linewidth]{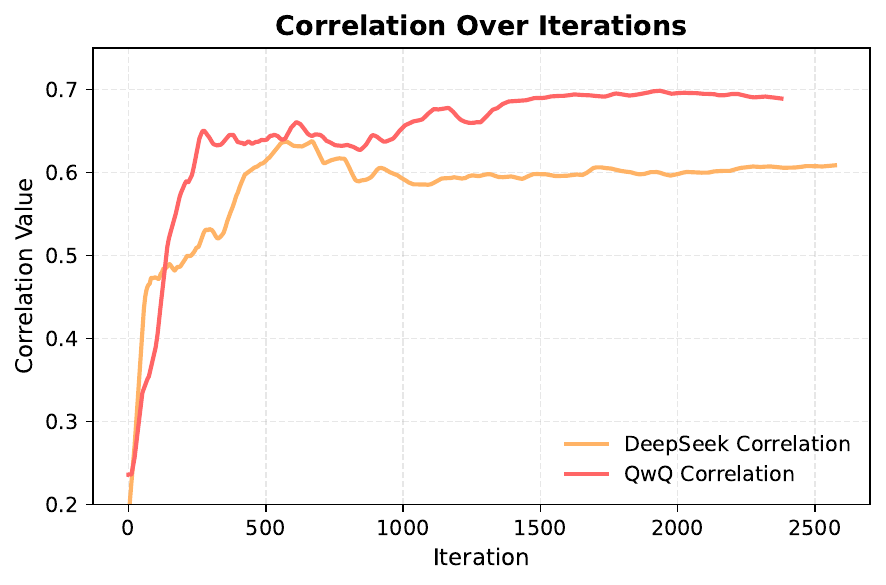}
    \caption{The Spearman Correlation over Pairwise Comparison Iterations.}
    \label{fig:corr_compare}
\end{figure}
\subsection{Results and Discussion}
\paragraph{Main Result} \autoref{tab::compare_result} and \autoref{fig:corr_compare} presents the results of probability estimation using different methods and models.
The final correlations from pairwise comparison process fall short of expectations and do not surpass the direct prompting method. We attribute this to the limited reliability of the pairwise comparison judgments, particularly due to the noisiness in comparing semantically unrelated NLI passages. 
While pairwise comparison is often considered cognitively easier in human evaluation settings, in our case, the LLMs require comparable amounts of reasoning and contextual understanding for both direct scoring and relative ranking. This undermines the presumed advantage of pairwise comparison in this task.
These results suggest that for probabilistic NLI tasks, pairwise comparison may not offer a significant benefit over direct prompting, especially when comparisons are performed over contextually disjoint samples. More structured or semantically clustered comparisons may be necessary to fully realize the benefits of pairwise reasoning.

\paragraph{Difficulty of NLI Comparison}
We employ GPT-4o~\citep{openai2024gpt4technicalreport} to do pairwise comparison. The comparison accuracy is evaluated against ground-truth pairwise labels, while entropy, computed from the API logits, serves as a measure of the model's uncertainty.

\begin{figure}[ht]
    \centering
    \begin{subfigure}[b]{0.48\textwidth}
        \centering
        \includegraphics[width=\textwidth]{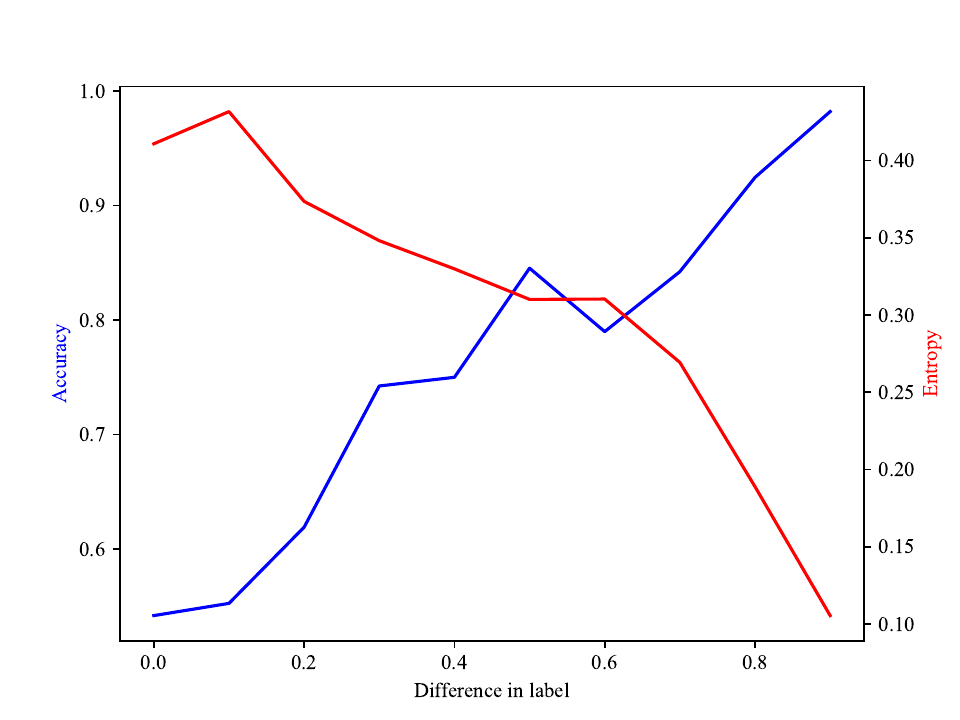}
        \caption{Accuracy and entropy of pairwise comparison as a function of label difference.}
        \label{fig:accuracy_entropy}
    \end{subfigure}
    \hfill
    \begin{subfigure}[b]{0.48\textwidth}
        \centering
        \includegraphics[width=\textwidth]{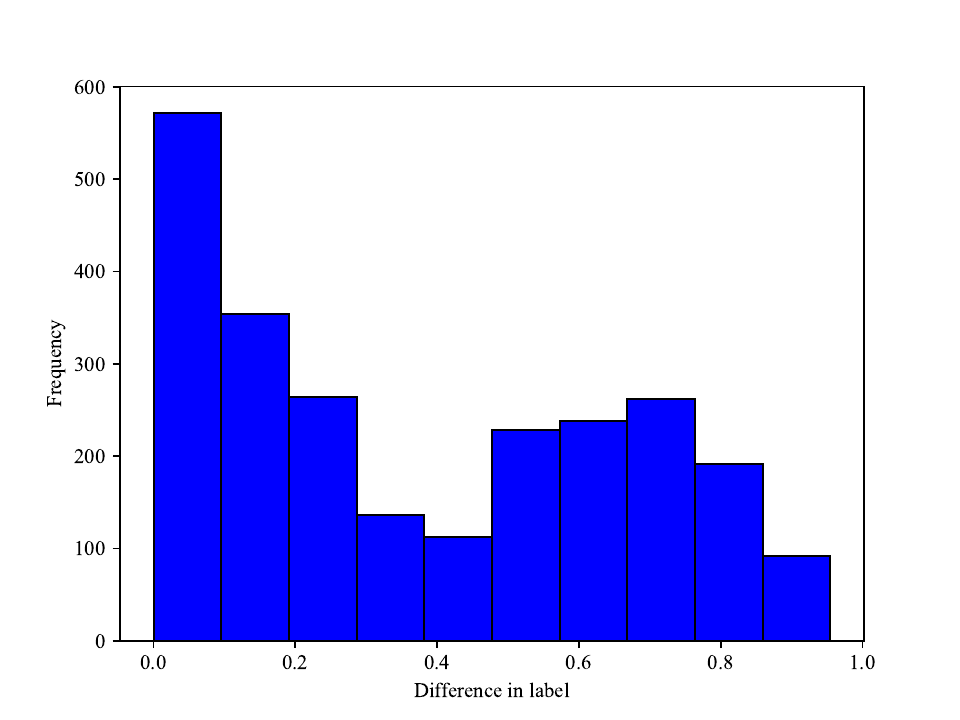} 
        \caption{Distribution of label differences in the sampled data.}
        \label{fig:label_distribution}
    \end{subfigure}
    \caption{Analysis of GPT-4.0 on NLI probability pairwise comparison tests. (a) Higher label differences lead to improved accuracy and reduced entropy. (b) Frequency distribution of label differences across the sampled data.}
    \label{fig:gpt4_comparison}
\end{figure}

As illustrated in Figure~\ref{fig:gpt4_comparison}, the trends in accuracy and entropy align with intuitive expectations. Specifically, as the difference in label probabilities increases, the comparison accuracy improves, and the entropy decreases. This pattern suggests that the model becomes more confident in its decisions when the distinction between the samples is more significant. The results highlight GPT-4o can produce reliable judgments in scenarios with greater label separability while reflecting higher uncertainty in ambiguous cases.

\section{Analysis of Loss Function Gradients}
\label{appendix::gradient}
 
We analyses the gradients of different losses we adopted to the class-wise probability after the softmax layer. We consider the following losses. Suppose that $b_i$ is the i-th bin, and $f(b_i)$ is the corresponding scalar score. Let $\mathbf{p}$ be the predictive distribution from the model and each $p_i$ be the predicted probability for the i-th bin, and $\bm{\pi}$ and $\pi_i$ the corresponding logits respectively. $s$ be the ground truth scalar score.

\paragraph{Mean Square Error} takes the form

\begin{align*}
    l_{\text{MSE}}(\bm{\pi}) &= \frac{1}{2}\Big(\sum_{i = 0}^{N - 1}p_i \cdot f(b_i) - s\Big)^2,\ \text{where the gradient is given by} \\
    \frac{\partial l_{\text{MSE}}}{\partial \pi_i} &= \Big(\sum_{i = 0}^{N - 1}p_i \cdot f(b_i) - s\Big)p_i\Big(f(b_i) - s\Big).
\end{align*}

Therefore, if the model underpredict the probability, the logits for all bins above the true value will be increased, and the more the differences between the bin value and the true value, the more the logits will be increased. This is not ideal as we typically want to model to center its prediction around the true value.

\paragraph{Marign Loss} For two instance $(x^j, s^j), (x^k, s^k)$, the uncropped margin loss takes the form

\begin{align*}
  l_{\text{Margin}}(\bm{\pi}^j, \bm{\pi}^k)& = - \text{sgn}(s^j - s^k)\cdot\Big(\sum_{i = 0}^{N - 1} p^j_i f(b_i) - \sum_{i = 0}^{N - 1} p^k_i f(b_i)\Big),\ \text{where}\\
  \frac{\partial l_{\text{Margin}}}{\partial \pi^j_i} &= \Big(f(b_i) - \sum_{i = 0}^{N - 1} p^j_i f(b_i)\Big)p^j_i,
\end{align*}

which similarly encourages the model to further updates the logits for the further away bins from the current prediction. To support greedy decoding of reasonable coarse labels, on human annotated data, we still train with the KL-divergence loss to match the quantized label distribution.

\section{Aligning with Human Crowd Distributions}
\label{appendix::align_with_human}
To evaluate the instance-level calibration of our model, we assess its ability to produce probability estimates that align with aggregated human judgments for the same instance. This is a crucial aspect of building trustworthy models whose confidence reflects human consensus. We conduct two experiments on datasets with rich human annotations to validate this capability.

First, we utilize the ChaosNLI dataset~\citep{nie-etal-2020-learn}, which provides 100-annotator voting distributions for NLI examples. Following a similar methodology to~\citet{jiang2024addressing}, we map the distributions onto a single scalar probability score to serve as the ground truth. Specifically, we use the mapping rule: $$p(\text{entailment}) = 1.0, p(\text{neutral}) = 0.2, p(\text{contradiction}) = 0.0.$$ As shown in \autoref{tab:chaosnli_results}, our model's predicted probabilities achieve a high Spearman's $\rho$ correlation with these human-derived scores and a correspondingly low ranking risk. This indicates a strong monotonic relationship between our model's confidence and the consensus of a human crowd.

\begin{table}[h!]
\centering
\small
\begin{tabular}{l c c c}
\toprule
\textbf{Dataset} & \textbf{Total Examples} & \textbf{Spearman's $\rho$ $\uparrow$} & \textbf{Ranking Risk $\downarrow$} \\
\midrule
ChaosNLI-S & 1514 & 0.9029 & 0.1959 \\
ChaosNLI-M & 1599 & 0.8031 & 0.1959 \\
\bottomrule
\end{tabular}
\caption{Correlation and ranking risk between our model's predictions and aggregated human judgments on ChaosNLI. High correlation and low risk demonstrate strong alignment.}
\label{tab:chaosnli_results}
\end{table}

To further validate these findings, we evaluate our model's calibration performance on the ProtoQA dataset~\citep{boratko-etal-2020-protoqa}, a commonsense question-answering task where each question has multiple plausible answers annotated with human vote frequencies. For each question-answer pair $(q, a_i)$, we treat our model's predicted probability, $p(a_i|q)$, as the output of a probabilistic classifier. Table~\ref{tab:protoqa_results} presents the standard calibration metrics. The low Expected Calibration Error (ECE), Brier Score, and Jensen-Shannon Divergence (JSD) demonstrate that the model's predictive distribution is well-calibrated against the distribution of human judgments.

\begin{table}[h!]
\centering
\small
\begin{tabular}{l c c c}
\toprule
\textbf{Metric} & ECE $\downarrow$ & Brier Score $\downarrow$ & JSD $\downarrow$ \\
\midrule
\textbf{Value} & 0.0105 & 0.0159 & 0.0499 \\
\bottomrule
\end{tabular}
\caption{Calibration metrics on ProtoQA. Low error values across all metrics confirm that the model's predictive distribution aligns well with human answer frequencies.}
\label{tab:protoqa_results}
\end{table}

Taken together, the high rank correlation on ChaosNLI and the low calibration error on ProtoQA strongly suggest that our model's probability estimates effectively mirror human-annotated label distributions.

\section{Prompt Template}
Here are the prompts used in our experiments, with input data inserted into the curly brackets.
\subsection{Probability Extraction}
\label{app_prompt}

\begin{lstlisting}
system: You are a helpful assistant good at probabilistic reasoning in the real world setting,
human: Given a premise and a hypothesis, evaluate the probability of the hypothesis being true based on the information provided in the premise, supplemented by world knowledge and probabilistic reasoning. 

Specifically:
1. Use relevant world knowledge to assess contextual factors (e.g., demographics, common practices, or statistical distributions) that may influence the likelihood of the hypothesis given the premise.
2. Perform the probabilistic reasoning to estimate the conditional probability P(Hypothesis | Premise).
3. Assign a probability score between [0, 1] that quantifies P(Hypothesis | Premise). Ensure this score reflects the strength of the connection between the premise and hypothesis based on probabilistic reasoning and world knowledge.

Premise: {premise}
Hypothesis: {hypothesis}
              
Your final probability estimate should be a value in the range [0,1], as fine-grained as possible, and formatted as follows: ```your_final_probability```.
For example, if the estimated probability is 0.0653, the output should be: ```0.0653```
\end{lstlisting}
\subsection{Reasoning Summary}
\begin{lstlisting}
system: You are a helpful assistant good at reasoning summary,
human: Simply **summarize** the reasoning process and final estimated probability (a float between 0 and 1). Include key factors, assumptions, and influences on the estimated probability, dismiss repetitive information.

Reasoning: {reasoning}

Begin your summary with "Reasoning:" and end with "Probability:".
\end{lstlisting}
\subsection{0-shot Probability Scoring}
\label{appendix::zeor-shot}

\begin{lstlisting}
system: You are a helpful assistant good at probabilistic reasoning in the real world setting.
human: Your task is to estimate the probability of a textual outcome $O$ given a description of the context $C$. Please respond with a probability in the range of [0, 1] that's your best estimate of the conditional probability  P(O|C).

### Cotext
{context}

### Outcome
{outcome}

Wrap your final answer in triple quotes format. Please don't generate any other text.
\end{lstlisting}

\subsection{Judge with Confidence}
\begin{lstlisting}
system: You are a helpful assistant good at probabilistic reasoning in the real world setting
human: Define the Random Variables:
- Let \( H \) represent the hypothesis: {hypothesis}
- Let \( P \) represent the premise: {premise}

Below are the estimation of conditional probability P(H|P) based on the given premise and hypothesis from other agents using probabilistic reasoning and real-world knowledge.
Your task is to assign a confidence score (ranging from 0 to 1) to each agent's response.
                 
1. {reasoning_1}
              
2. {reasoning_2}
              
3. {reasoning_3}

4. {reasoning_4}

Important Considerations: 
- Think like a human, go beyond literal semantics by considering context, common sense, and real-world knowledge.
- Related premise and hypothesis do not necessarily cause high probability.
- Assign higher confidence to assumptions that are more commonly observed and reasoning processes that are logically sound, fully justified.
- If the premise and hypothesis both refer to an entity using an indefinite noun phrase (e.g., 'a person'), assume they refer to the same entity unless there is clear evidence suggesting otherwise.
- Errors may arise from disagreements in reasoning, differing assumptions, logical mistakes, overemphasis on corner cases, underconfidence or overconfidence, and inaccuracies in estimating P(H|P). Do not blindly trust other agents' probability estimation.

Here are some examples of probability extimation:
1. hypothesis: Three brothers pound on some drums
   premise: Three men dressed in white shirts and white hats, (two with baseball caps, the leader with a white construction helmet), pounding sticks on steel and plastic drums. 
   probability: 0.00000027      
2. hypothesis: There is a rock currently skipping down a pond.
   premise: A young african boy skipping rocks.
   probability: 0.058
3. hypothesis: The man is walking into a room.
   premise: A man is standing in the doorway of a building.
   probability: 0.2639
4. hypothesis: People are rollerblading for something to do.
   premise: At least six individuals are on a team wearing helmets and knee pads while rollerblading around a skating rink.
   probability: 0.5
5. hypothesis: A brown dog is outside and it's snowing
   premise: A brown dog plays in a deep pile of snow.
   probability: 0.7342
6. hypothesis: Two girls attend a convention.
   premise: Two girls in a crowd are dressed up, one as the cartoon character Wall-E.
   probability: 0.94
7. hypothesis: Some kids splash in the water and interact with each other.
   premise: many children play in the water.
   probability: 0.99
              
Output Format:
- The confidence score for other agents should be a decimal value between 0 and 1, formatted as: \\boxed{{confidence1, confidence2,confidence3,confidence4}}
- Example output: \\boxed{{0.1,0.5,0.8,0.2}}
\end{lstlisting}
\subsection{Pairwise Compare}
\begin{lstlisting}
system: You are a helpful assistant,
human: Given the premise and hypothesis of two natural inference passages, determine which hypothesis is more likely based on its premise.

Specifically:
1. Contextual Assessment with World Knowledge
Analyze each pair: Evaluate the premise and hypothesis using relevant world knowledge. Consider contextual factors such as demographics, common practices, or statistical distributions to estimate the likelihood of the hypothesis being true. State assumptions: Explicitly identify any assumptions or uncertainties introduced by missing information in the premise.
2. Comparison
Compare the likelihood of each hypothesis based on the alignment between the premise and hypothesis.
Justify your reasoning for why one hypothesis is more likely than the other, considering the degree of alignment and the assumptions made.
If the likelihoods of both hypotheses are sufficiently close or indistinguishable, return a None.

Passage A: {item1}
Passage B: {item2}

3. Output Format Example: 
In your final decision, strictly output \boxed{{Passage A}}, \boxed{{Passage B}} or \boxed{{None}}")])
\end{lstlisting}

\subsection{EntailmentBank, e-CARE Rewrite}
\begin{lstlisting}
system: You are a helpful assistant"),
human: Given a natural language inferenc passage: {passage}
Your goal:
Rewrite the original premise and hypothesis 
Generate 2 new premises related to the passage so that the probability of NLI P(hypothesis | new premise1, new premise2) ranges from **0.05 to 0.95**.

Steps to follow:
1. Rewrite the original premise and hypothesis for clarity and precision
- Ensure both the premise and hypothesis are clear, precise, and logically sound.
- Removed unnecessary modal verbs (e.g., "can," "might") and hedging language (e.g., "possibly," "somewhat").
- If needed, specify a concrete example for clarity.
2. Generate new premises that modify the likelihood of the hypothesis being inferred:
- Ensure all generated premises are factually correct and logically consistent.
- Here are strategies you may consider to adjust the probability of inference:
- Alternative Explanation (Misattribution): Provide a different cause for the phenomenon, weakening or shifting inference.
- Increase Vagueness: Make the premise more general, requiring additional inference.
- Observer-Dependent Effects: Frame the premise in a way that makes the inference more subjective or situational.
- Instantiation: Provide a specific example that either supports or challenges the inference.
3. Categorize Premises into Four Bins Based on Probability
- highly likely(probability~0.9): Premises that strongly support the hypothesis but may introduce slight variation or broader interpretations.
- moderately likely(probability~0.7): Premises that are related to the passage but are more general, potentially requiring additional context to confirm the hypothesis.
- neutral(probability~0.5): Balances support and doubt, making the inference uncertain.
- unlikely (probability~0.3): Premises that introduce alternative mechanisms or are only tangentially related to the hypothesis.
- contradict(probability~0.1): Challenges the hypothesis by shifting the explanation, observer perspective, or causal factor without introducing factual errors.
4. Format the output as a valid JSON object with the following structure:
{{
  "premise": "Your revised premise here.",
  "hypothesis": "Your revised hypothesis here.",
  "highly likely": [premise1, premise2],
  "moderately likely": [premise1, premise2],
  "neutral": [premise1, premise2],
  "unlikely": [premise1, premise2],
  "contradict": [premise1, premise2]
}}
5. Recheck that output statements are factually accurate and format is a valid JSON.
\end{lstlisting}

\newpage

\section{Example Structural Reasoning Traces}
\label{appendix::mermaid}

In this section, we provide example traces constructed to evaluate local scoring models. The example reasoning trace corresponds to an instance from the BIRD dataset for C2S-Sent-B \citep{feng2024birdtrustworthybayesianinference}. The block on the left describe a situation with an additional sentence supporting an outcome, and the block on the right describes potential outcomes. In this case BIRD infers the outcome the additional sentence support by reasoning over a range of factors and conditions probabilistically. The scores produced by our model are shown in gray blocks. Note that we do not perform NLI-based edge filtering, and the scores are not coherent.

    \centering
    \includegraphics[width=\linewidth]{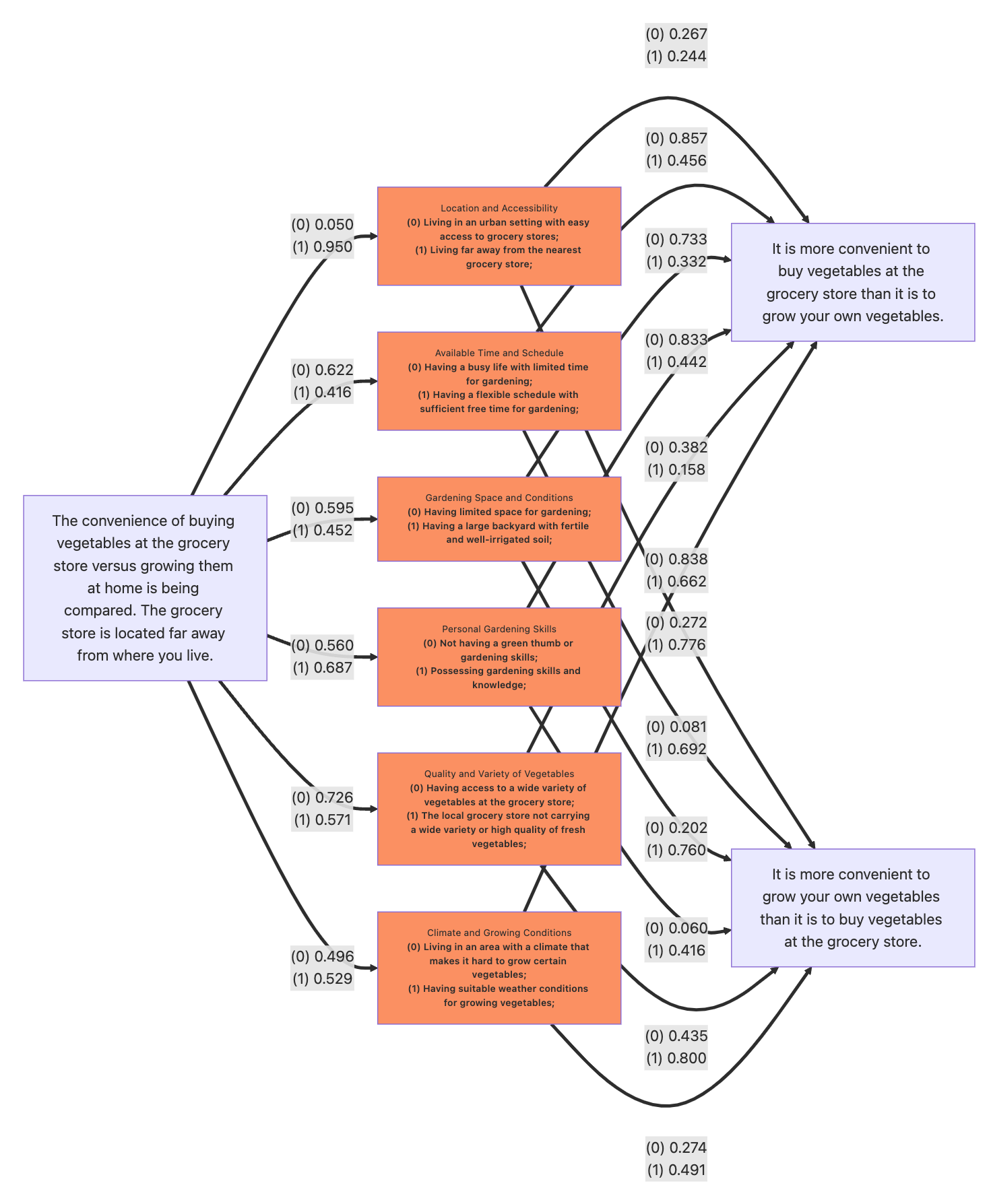}
\end{document}